\documentclass{ifacconf}

\makeatletter
\let\old@ssect\@ssect 
\makeatother

\usepackage{natbib}        
\usepackage{hyperref}

\makeatletter
\def\@ssect#1#2#3#4#5#6{%
  \NR@gettitle{#6}
  \old@ssect{#1}{#2}{#3}{#4}{#5}{#6}
}
\makeatother

\usepackage{url}
\usepackage{hyperref}
\usepackage{graphicx}      
\usepackage{natbib}        
\usepackage[export]{adjustbox}
\usepackage{comment}
\usepackage{color}
\usepackage{latexsym}
\usepackage{xurl}
\usepackage{mathtools}
\usepackage{caption}
\usepackage{subcaption}
\usepackage{amsmath,amssymb,amsfonts}
\usepackage{mathrsfs}
\usepackage{algorithmic}
\usepackage{pifont}
\usepackage{textcomp}
\usepackage{multirow}
\usepackage[table,xcdraw]{xcolor}
\usepackage{balance}

\DeclareMathAlphabet{\pazocal}{OMS}{zplm}{m}{n}

\begin{document}
\begin{frontmatter}

\title{A Systematic Digital Engineering Approach to Verification \& Validation of Autonomous Ground Vehicles in Off-Road Environments} 

\author[First]{Tanmay Samak$^1$}
\author[First]{Chinmay Samak$^1$}
\author[Second]{Julia Brault}
\author[Second]{Cori Harber}
\author[Second]{Kirsten McCane}
\author[Third]{Jonathon Smereka}
\author[Third]{Mark Brudnak}
\author[Third]{David Gorsich}
\author[First]{Venkat Krovi}
\thanks[authorinfo]{These authors contributed equally.}
\thanks[opsecinfo]{DISTRIBUTION STATEMENT A. Approved for public release; distribution is unlimited. OPSEC9523.}
\address[First]{CU-ICAR, Greenville, SC 29607, USA.}
\address[Second]{The MathWorks, Inc., Natick, MA 01760, USA.}
\address[Third]{U.S. Army DEVCOM GVSC, Detroit, MI 48092, USA.}

\begin{abstract} 
The engineering community currently encounters significant challenges in the systematic development and validation of autonomy algorithms for off-road ground vehicles. These challenges are posed by unusually high test parameters and algorithmic variants. In order to address these pain points, this work presents an optimized digital engineering framework that tightly couples digital twin simulations with model-based systems engineering (MBSE) and model-based design (MBD) workflows. The efficacy of the proposed framework is demonstrated through an end-to-end case study of an autonomous light tactical vehicle (LTV) performing visual servoing to drive along a dirt road and reacting to any obstacles or environmental changes. The presented methodology allows for traceable requirements engineering, efficient variant management, granular parameter sweep setup, systematic test-case definition, and automated execution of the simulations. The candidate off-road autonomy algorithm is evaluated for satisfying requirements against a battery of 128 test cases, which is procedurally generated based on the test parameters (times of the day and weather conditions) and algorithmic variants (perception, planning, and control sub-systems). Finally, the test results and key performance indicators are logged, and the test report is generated automatically. This then allows for manual as well as automated data analysis with traceability and tractability across the digital thread.
\end{abstract}

\begin{keyword} 
Autonomous Vehicles; Digital Twins; Digital Engineering; Model-Based Systems Engineering; Model-Based Design, Verification \& Validation
\end{keyword}

\end{frontmatter}


\section{Introduction}
\label{Section: Introduction}

\begin{figure*}[t]
\centering
\includegraphics[width=\linewidth]{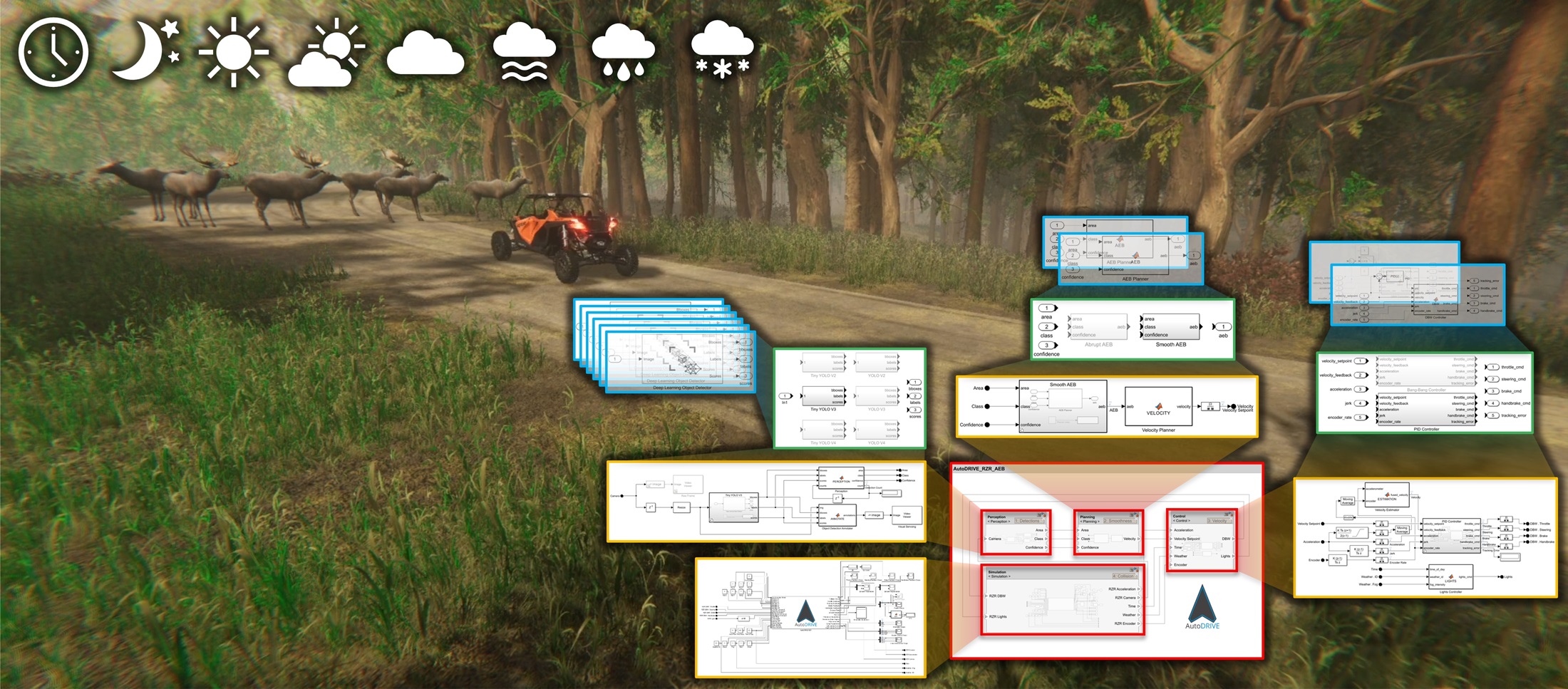}
\caption{Overview of the proposed framework depicting automated off-road autonomy validation using digital twins coupled with MBSE and MBD workflows. Video: \url{https://youtu.be/FsSTWJiiEWg}}
\label{fig1}
\end{figure*}

Autonomous vehicles (AVs) are complex, multidisciplinary systems that integrate mechanical, electrical, electronic, computing, software, and information sub-systems. The design and development of AVs are therefore critical challenges, often involving design choice(s) of certain components and sub-systems over other variants, with trade-offs that can impact system performance and safety. Additionally, given their safety-critical nature \cite{7823109}, AVs must undergo rigorous verification and validation (V\&V) processes \cite{10.1145/3542945, 7795548} to ensure their reliability across a variety of operating conditions. Capturing the right set of test parameters effectively for these processes presents another major challenge \cite{75eb35f0-ef72-3076-a9bc-01464345aecc}.

While vehicular autonomy has been explored over the past few decades, most of the existing approaches to designing, developing, and validating these systems have been relatively ad-hoc. Off-road autonomy further aggravates these problems by presenting unique challenges in terms of the optimal choice of algorithmic variants (e.g., different approaches to handle perception, planning, control, etc.) and unusually high test conditions (e.g., rough terrains, unpredictable lighting and weather conditions, varied types of obstacles, etc.). As such, developing a systematic approach that couples development and testing phases recursively, with automated decision-making, is essential. This would help identify potential risks early in the development process and ensure consistent quality throughout the lifecycle of autonomous systems.

In such a milieu, digital engineering \cite{doi:10.1177/1548512917747050, IAKSCH2019100942, dod-de} stands out as a promising workflow to standardize the development and testing of autonomous systems. Digital engineering refers to the integrated use of models, simulations, and data to support the lifecycle management of complex systems from conceptualization through disposal. It binds data-driven and model-based approaches into a unified framework, which creates a continuum across various engineering disciplines. Digital engineering can also leverage software tools to establish automated workflows that link requirements \cite{10.1145/3477314.3507004}, design, development, and testing, thereby reducing the reliance on manual intervention and making it possible to systematically explore and analyze numerous design choices in real-time.

This research proposes a specialized digital engineering framework that integrates digital twin simulations with model-based systems engineering (MBSE) and model-based design (MBD) paradigms for developing and validating off-road autonomous vehicles (refer Fig. \ref{fig1}). Here, MBSE provides a structured framework for modeling and analyzing the overall system architecture and requirements, helping engineers identify potential design flaws at the conceptual stage. MBD implements hierarchically consistent digital models of the components, sub-systems, systems, and potentially system-of-systems, thereby enabling recursive simulation, validation, and optimization. Digital twins enhance this process by creating physically and graphically accurate virtual representations of real-world systems, enabling real-time interfacing for monitoring and control. Together, the synergy between digital twins, MBSE, and MBD workflows promotes continuous integration and testing throughout the system's lifecycle, with tractability and traceability across the digital thread \cite{doi:10.2514/1.J057255}.

The key contributions of this work can be summarized as follows:

\begin{itemize}
    \item Developing high-fidelity digital twins of autonomous Polaris RZR Pro R 4 Ultimate vehicle and its operating off-road environment within the open-source AutoDRIVE Ecosystem\footnote{\url{https://autodrive-ecosystem.github.io}} \cite{AutoDRIVEEcosystem2023}.
    \item Integrating the digital twin framework with MBSE and MBD workflows to recursively design and test off-road autonomy algorithms by developing an open interface for MathWorks\textsuperscript{\textregistered} System Composer\textsuperscript{\texttrademark}\footnote{\url{https://www.mathworks.com/products/system-composer.html}}.
    \item Demonstrating the efficacy of the proposed framework through an end-to-end representative case study covering the granular stages from conceptualization to validation, including automated execution and decision-making.
\end{itemize}

The remainder of this paper is structured as follows: Section \ref{Section: Related Work} presents a literature survey of digital engineering tools and methodologies applied to autonomous systems. Section \ref{Section: Digital Engineering Framework} elucidates various elements of the proposed digital engineering framework and the systematic workflow. Section \ref{Section: Case Study} describes the design, development, verification, and validation of an off-road autonomy case study using the proposed framework. Finally, Section \ref{Section: Conclusion} provides concluding remarks and points to future research directions. 


\section{Related Work}
\label{Section: Related Work}

This section presents a systematic literature review of various digital engineering tools and methodologies that have been applied to the design, development, validation, and optimization of autonomous systems, and also highlights potential research gaps.

\subsection{Digital Twin Simulation}
\label{Sub-Section: Digital Twin Simulation}

Modeling and simulation \cite{9605690} are paramount to digital engineering. Industrial tools such as NVIDIA DRIVE Constellation \cite{DRIVEConstellation2019}, NVIDIA Isaac Sim \cite{IsaacSim2025}, dSPACE ADAS \& AD Portfolio \cite{dSPACE2021}, Simcenter Prescan \cite{PreScan2025}, and Hexagon VTD \cite{VTD2025}, among others, provide specialized solutions to designing and testing autonomous systems that are often compliant with industry standards and regulations. However, most of these tools are propitiatory, allowing limited parameter tuning and reconfiguration for modeling, simulation, and test case generation. Contrarily, open-source simulators such as Gazebo \cite{Gazebo2004}, CARLA \cite{CARLA2017}, AirSim \cite{AirSim2018}, and LGSVL Simulator \cite{LGSVLSimulator2020}, among others, offer open-interfaces to customize simulations flexibly, but may lack precise/calibrated real-world counterparts, rendering them unsuitable for ``digital twinning'' applications.

\subsection{Model-Based Systems Engineering}
\label{Sub-Section: Model-Based Systems Engineering}

MBSE is an integral part of digital engineering, which focuses on managing the overall system of systems with hierarchical consistency. Although there is a vast amount of literature about MBSE in general and its applications to various fields, here, we point to some of the contributions specifically in the field of automotive engineering \cite{8122925}. \cite{VeeramaniLekamani1335887} explores the application of SysML \cite{sysml, dod-sysml} for autonomous vehicle trajectory planning and elucidates the workflow of defining the system architecture, specifying requirements, and performing analysis of the system functions. \cite{10417219} apply MBSE to advanced driver assistance system (ADAS) development and highlight the potential benefit of its integration with scenario generation methods to ensure comprehensive testing of system requirements. \cite{systems7010001} explore the application of MBSE to develop a resilient lane-centering function for automotive systems. \cite{10660657} apply MBSE to design an emergency braking function for automobiles incorporating security patterns into the system model, emphasizing the inclusion of cyber-security and functional safety. \cite{systems7010007} present a high-level vision and rationale for incorporating digital twin technology into MBSE.

\subsection{Model-Based Design}
\label{Sub-Section: Model-Based Design}

MBD contributes to digital engineering by modeling and simulating individual components, sub-systems, and systems within the larger system-of-systems hierarchy. \cite{7419080} present model-based design and validation of an autonomous vehicle using the bond graph method. \cite{9591251} explore MBD for model-based trajectory tracking control of an omnidirectional automated guided vehicle (AGV). \cite{10196233} suggest a formal extension of the V-model \cite{GAUSEMEIER2002785} describing the mechatronics approach of system design, verification, and validation for autonomous vehicles. \cite{9304603} leverage MBD for scenario-based testing of autonomous vehicles, particularly demonstrating a case study of emergency braking application in response to an occluded pedestrian potentially crossing the road. \cite{8226993} explore MBD for supervisory control of low-level ADAS functions such as cruise control and adaptive cruise control. \cite{YU2024803} apply MBD to design a linear model predictive controller for automated steering control of articulated vehicles such as tractor-trailers.

\subsection{Research Gaps}
\label{Sub-Section: Research Gaps}

Despite recent studies exploring digital twins, MBSE, and MBD, several research gaps remain that require attention. Firstly, most existing simulators target indoor or on-road autonomy, with no specific focus on off-road autonomy. Secondly, the community lacks a comprehensive framework that tightly integrates digital twins with MBSE and MBD workflows, which could foster more accessible, scalable, and interoperable systems. Finally, although there are numerous case studies in autonomous driving, end-to-end case studies for off-road vehicles -- covering systematic stages from \textit{``cradle-to-grave''} -- are less well-studied.


\section{Digital Engineering Framework}
\label{Section: Digital Engineering Framework}

This section highlights the proposed digital engineering framework for conceptualizing, designing, developing, integrating, and validating off-road autonomy algorithms. The framework couples the high-fidelity digital twin simulations of AutoDRIVE Simulator \cite{AutoDRIVESimulator2021} with an integrated model-based engineering (MBSE + MBD) approach in MathWorks\textsuperscript{\textregistered} System Composer\textsuperscript{\texttrademark}, which allows systematic development and assessment of off-road autonomy algorithms.

\subsection{Digital Twins}
\label{Sub-Section: Digital Twins}

Digital engineering relies heavily on mathematical or computational models of systems, which serve as the \textit{``single source of truth'' (SSOT)} for design, analysis, and optimization. Consequently, it stands to gain significantly from the promising capabilities of digital twins, which allow for the virtual replication of physical systems and enable seamless reality-to-simulation (real2sim) and simulation-to-reality (sim2real) updates throughout the system's lifecycle.

\begin{figure}[htpb]
\centering
\includegraphics[width=\linewidth]{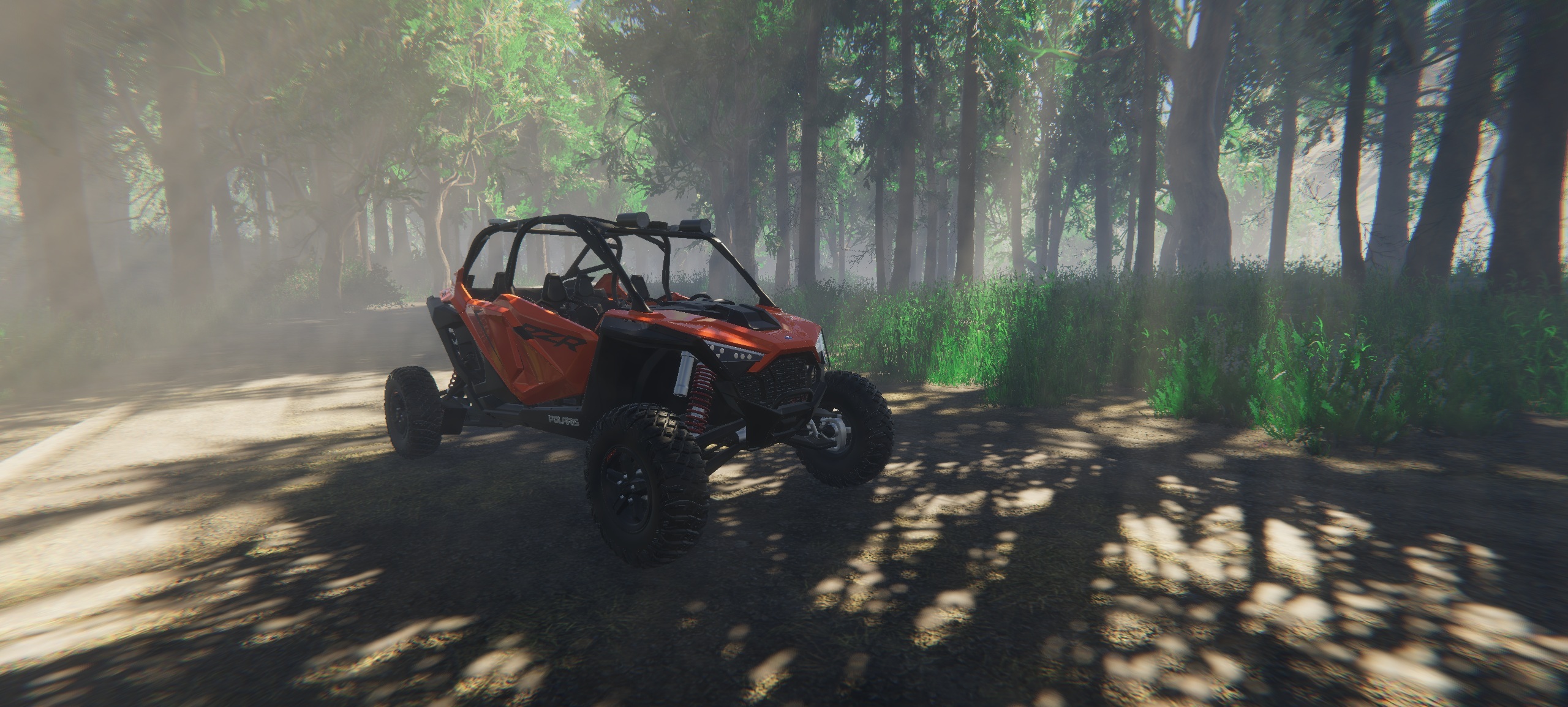}
\caption{Digital twin of the autonomous Polaris RZR Pro R 4 Ultimate within a virtual forest environment.}
\label{fig2}
\end{figure}

\subsubsection{Vehicle Simulation}
\label{Sub-Sub-Section: Vehicle Simulation}

\begin{figure}[t]
\centering
\includegraphics[width=\linewidth]{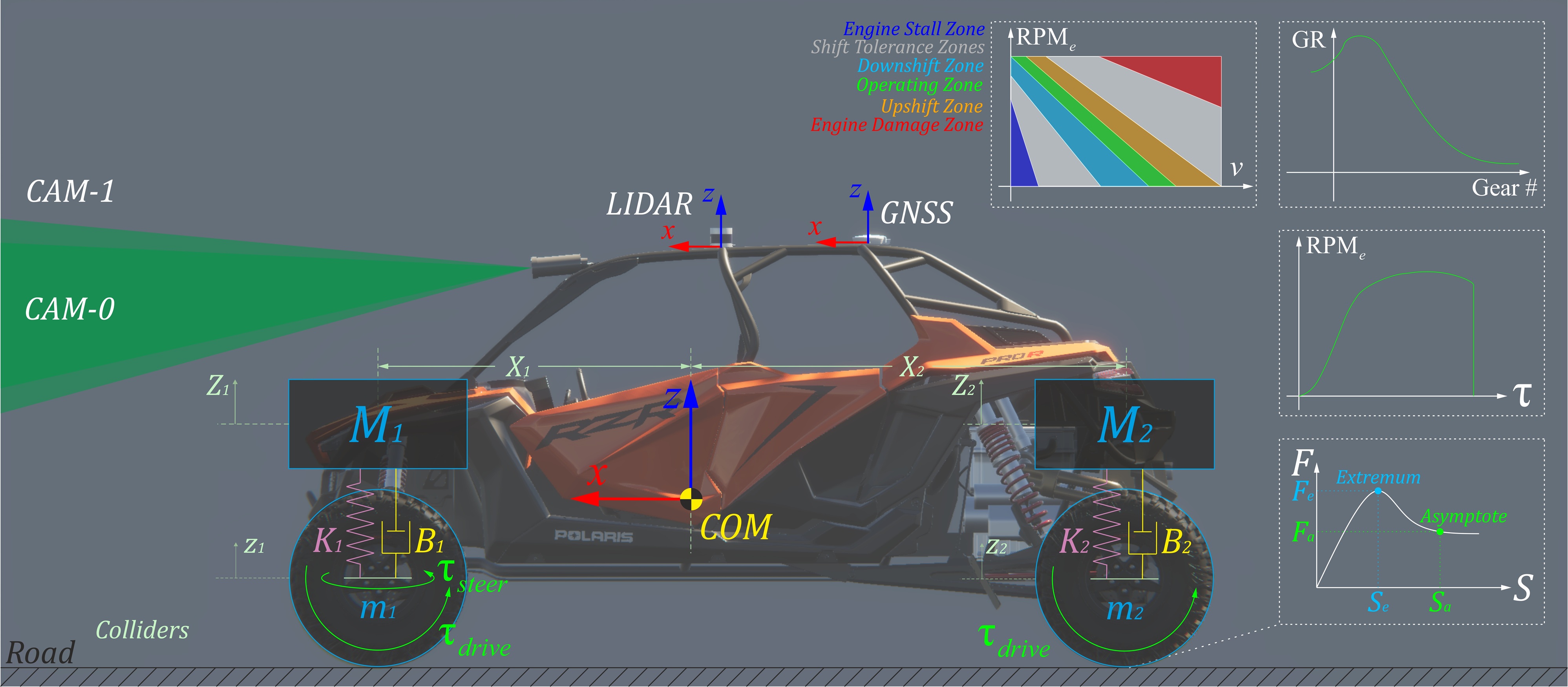}
\caption{Simplified representation of the vehicle dynamics and sensor simulation models of the RZR digital twin.}
\label{fig3}
\end{figure}

The vehicle (refer Fig. \ref{fig3}) is jointly represented as a sprung-mass ${^iM}$ with rigid-body dynamics. The two representations are linked by the total mass $M=\sum{^iM}$, center of mass, $X_\text{COM} = \frac{\sum{{^iM}*{^iX}}}{\sum{^iM}}$ and moment of inertia $I_\text{COM} = \sum{{^iM}*{^iX^2}}$. The wheels are also modeled as rigid bodies with mass $m$, subject to gravitational and suspension forces: ${^im} * {^i{\ddot{z}}} + {^iB} * ({^i{\dot{z}}}-{^i{\dot{Z}}}) + {^iK} * ({^i{z}}-{^i{Z}})$. Here, the stiffness ${^iK} = {^iM} * {^i\omega_n}^2$ and damping $^iB = 2 * ^i\zeta * \sqrt{{^iK} * {^iM}}$ of the suspension system are determined by the sprung mass ${^iM}$, natural frequency ${^i\omega_n}$, and damping ratio ${^i\zeta}$.

The vehicle's powertrain generates a torque equivalent to $\tau_{\text{total}} = \left[\tau_e\right]_{RPM_e} * \left[GR\right]_{G_\#} * FDR * \tau * \mathscr{A}$, were, $\tau_e$ is the engine torque, $\tau$ is the throttle input, and $\mathscr{A}$ is a non-linear smoothing operator. The engine speed smoothly updates as $RPM_e := \left[RPM_i + \left(|RPM_w| * FDR * GR\right)\right]_{(RPM_e,v)}$ where, $RPM_i$ is the idle speed, $RPM_w$ is the mean wheel speed, $FDR$ is the final drive ratio, $GR$ is the current gear ratio, and $v$ is the vehicle velocity. It is kept in check by the automatic transmission, which maintains a good operating range for a given speed: $RPM_e = \frac{{v_{\text{MPH}} * 5280 * 12}}{{60 * 2 * \pi * R_{\text{tire}}}} * FDR * GR$. The output torque at wheels depends on the vehicle's drive configuration:
$
\tau_{\text{out}} = \begin{cases}
\frac{\tau_{\text{total}}}{2} & \text{if FWD/RWD} \\
\frac{\tau_{\text{total}}}{4} & \text{if AWD}
\end{cases}
$.
The actual torque transmitted to the wheels $\tau_w$ is modeled by dividing the output torque $\tau_\text{out}$ to the left $^{L}\tau_{w} = \tau_{\text{out}} * (1 - \tau_{\text{drop}} * |\delta^{-}|)$ and right $^{R}\tau_{w} = \tau_{\text{out}} * (1 - \tau_{\text{drop}} * |\delta^{+}|)$ wheels based on the steering input. Here, $\tau_\text{drop}$ is the torque drop at differential and $(\tau_{\text{drop}} * |\delta^{\pm}|)$ is clamped between $[0,0.9]$.

The vehicle's braking torque is represented by ${^i\tau_\text{brake}} = \frac{{^iM}*v^2}{2*D_\text{brake}}*R_b$, where $R_b$ is the brake disk radius and $D_\text{brake}$ is the braking distance, calibrated at 60 MPH.

The vehicle's front wheels are steered based on the Ackermann steering geometry, considering $l$ as the wheelbase and $w$ as the track width:
$ \delta_{l/r} = \textup{tan}^{-1}\left(\frac{2*l*\textup{tan}(\delta)}{2*l\pm w*\textup{tan}(\delta)}\right)$.
Here, the steering rate is governed by $\dot{\delta} = \kappa_\delta + \kappa_v * \frac{v}{v_\text{max}}$, where $\kappa_\delta$ is the steering sensitivity and $\kappa_v$ is the speed-dependency factor of the steering mechanism.

The road-tire interconnect is simulated by modeling the friction curve for each tire, represented as $\left\{\begin{matrix} {^iF_{t_x}} = F(^iS_x) \\{^iF_{t_y}} = F(^iS_y) \\ \end{matrix}\right.$, where $^iS_x$ and $^iS_y$ denote the longitudinal and lateral slip values of the $i$-th tire. Here, the friction curve is modeled by a two-piece cubic spline $F(S) = \left\{\begin{matrix} f_0(S); \;\; S_0 \leq S < S_e \\ f_1(S); \;\; S_e \leq S < S_a \\ \end{matrix}\right.$, with $f_k(S) = a_k*S^3+b_k*S^2+c_k*S+d_k$. The first segment of the spline spans from the origin $(S_0,F_0)$ to the extremum $(S_e,F_e)$, while the second segment ranges from the extremum $(S_e, F_e)$ to the asymptote $(S_a, F_a)$.

The vehicle-environment interaction also accounts for variable air drag $F_\text{aero}$, based on the operating condition:

$
\begin{cases}
F_{d_\text{max}} & \text{if } v \geq v_{\text{max}} \\
F_{d_\text{idle}} & \text{if } \tau_{\text{out}} = 0 \\
F_{d_\text{rev}} & \text{if } (v \geq v_{\text{rev}}) \land (G_\# = -1) \land (RPM_{w} < 0) \\
F_{d_\text{idle}} & \text{otherwise}
\end{cases}
$

Here, $v$ is the velocity, $v_\text{max}$ is the practical top-speed, $v_\text{rev}$ is the practically maximum reverse velocity, $G_\#$ is the operating gear, and $RPM_w$ is the mean wheel speed.

\subsubsection{Sensor Simulation}
\label{Sub-Sub-Section: Sensor Simulation}

The autonomy-oriented digital twin of the RZR hosts a variety of proprioceptive and exteroceptive sensors for state estimation and scene perception. These sensors are modeled and characterized to mimic their real-world counterparts.

The simulated vehicle provides real-time feedback from the throttle ($\tau$), steering ($\delta$), brake ($\chi$), and parking brake ($\xi$) actuators. Additionally, it hosts incremental encoders to measure $^iN_{\text{ticks}} = {^iPPR} * {^iCGR} * {^iN_{\text{rev}}}$, where $^iPPR$ is the encoder resolution, $^iCGR$ is the cumulative gear ratio, and $^iN_{\text{rev}}$ represents the wheel revolutions.

The virtual inertial navigation system (INS) comprises GNSS and IMU, which are simulated based on temporally coherent rigid-body transform updates of the vehicle $\{v\}$ w.r.t. world $\{w\}$: ${^w\mathbf{T}_v} = \left[\begin{array}{c | c} \mathbf{R}_{3 \times 3} & \mathbf{t}_{3 \times 1} \\ \hline \mathbf{0}_{1 \times 3} & 1 \end{array}\right] \in SE(3)$.

The virtual cameras are simulated using a standard viewport rendering pipeline. First, the camera view matrix $\mathbf{V} \in SE(3)$ is computed by obtaining the relative homogeneous transform of the camera $\{c\}$ w.r.t. world $\{w\}$. Next, the camera projection matrix $\mathbf{P} \in \mathbb{R}^{4 \times 4}$ is calculated to project world coordinates into image space coordinates: $\mathbf{P} = \begin{bmatrix} \frac{2*N}{R-L} & 0 & \frac{R+L}{R-L} & 0 \\ 0 & \frac{2*N}{T-B} & \frac{T+B}{T-B} & 0 \\ 0 & 0 & -\frac{F+N}{F-N} & -\frac{2*F*N}{F-N} \\ 0 & 0 & -1 & 0 \\ \end{bmatrix}$, where $N$ and $F$ are the near and far clipping planes of the camera, while $L$, $R$, $T$, and $B$ denote the left, right, top, and bottom offsets of the sensor. Finally, a post-processing step replicates physical phenomena like lens distortion, depth of field, exposure, ambient occlusion, contact shadows, bloom, motion blur, film grain, chromatic aberration, and other lens and film effects.

The virtual 3D LIDAR is simulated by performing parallelized multi-channel ray-casting: \texttt{raycast}\{$^w\mathbf{T}_l$, $\vec{\mathbf{R}}$, $r_{\text{max}}$\} for each angle $\theta \in \left [ \theta_{\text{min}}:\theta_{\text{res}}:\theta_{\text{max}} \right ]$ and each channel $\phi \in \left [ \phi_{\text{min}}:\phi_{\text{res}}:\phi_{\text{max}} \right ]$ at a specified update rate, with GPU acceleration (if available). Here, ${^w\mathbf{T}_l} = {^w\mathbf{T}_v} * {^v\mathbf{T}_l} \in SE(3)$ represents the relative transformation of the LIDAR \{$l$\} w.r.t. the vehicle \{$v$\} and the world \{$w$\}, $\vec{\mathbf{R}} = \left [\cos(\theta)*\cos(\phi) \;\; \sin(\theta)*\cos(\phi) \;\; -\sin(\phi) \right ]^T$ defines the direction vector of each ray-cast $R$, where $r_{\text{min}}$ and $r_{\text{max}}$ denote the minimum and maximum linear ranges, $\theta_{\text{min}}$ and $\theta_{\text{max}}$ denote the minimum and maximum horizontal angular ranges, $\phi_{\text{min}}$ and $\phi_{\text{max}}$ denote the minimum and maximum vertical angular ranges, and $\theta_{\text{res}}$ and $\phi_{\text{res}}$ represent the horizontal and vertical angular resolutions of the LIDAR, respectively. The thresholded ray-cast hits $\{\mathbf{x_{\text{hit}}, y_{\text{hit}}, z_{\text{hit}}}\}$ are returned as point cloud data (PCD).

\subsubsection{Environment Simulation}
\label{Sub-Sub-Section: Environment Simulation}

\begin{figure}[h]
\centering
\includegraphics[width=\linewidth]{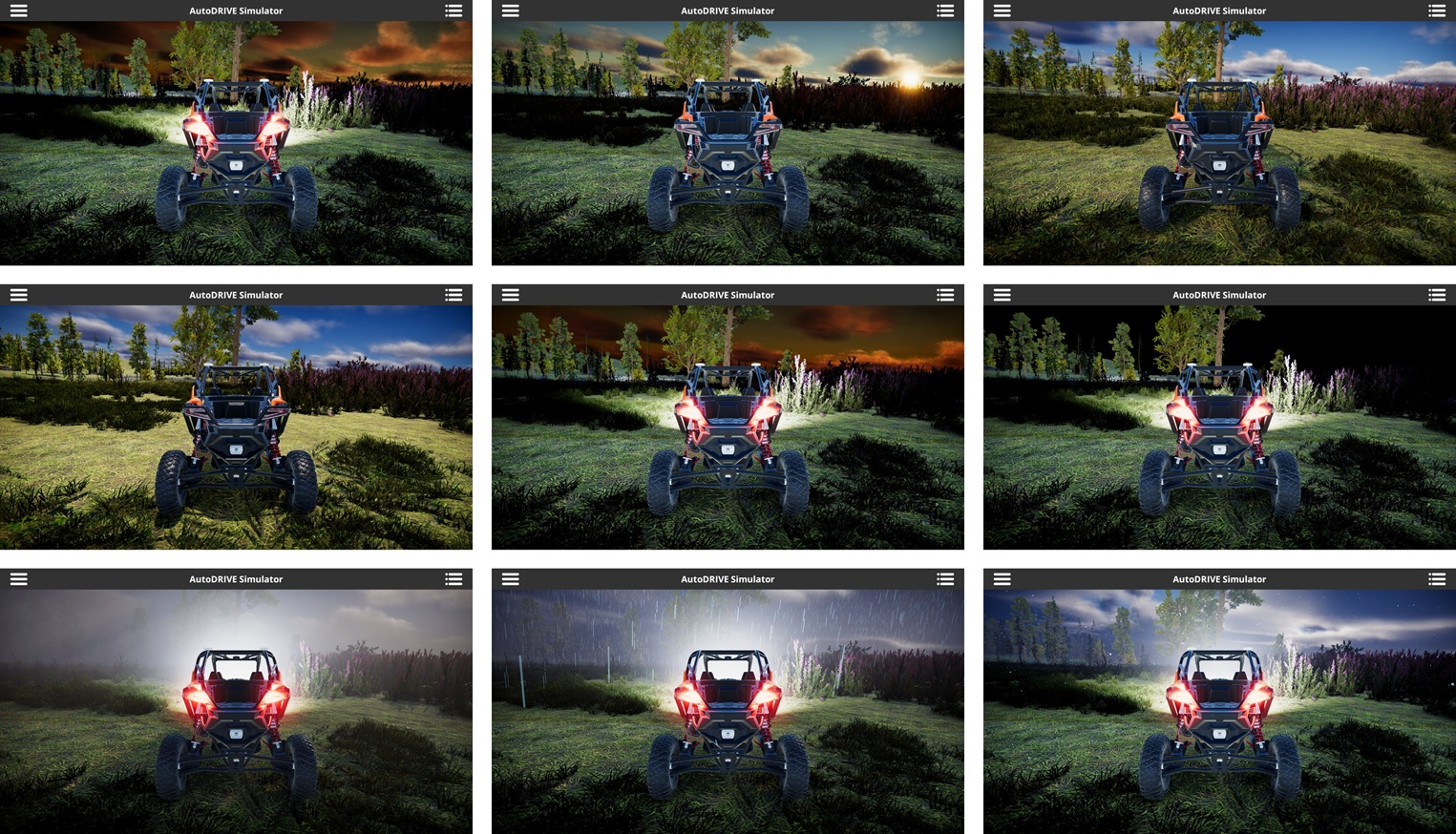}
\caption{Dynamic simulation of the virtual forest environment across different times of the day (top 2 rows depict times from dawn to night) and weather conditions (the last row depicts fog, rain, and snow).}
\label{fig4}
\end{figure}

The static environment (refer Fig. \ref{fig2}) is simulated by performing mesh-mesh interference detection and calculating contact forces, frictional forces, momentum transfer, as well as linear and angular drag exerted on all rigid bodies. Additionally, the simulator can introduce dynamic variability in terms of changing the time of day and weather conditions (refer Fig. \ref{fig4}). Particularly, the simulator models physically-based sky and celestial bodies, enabling the simulation of sunlight and moonlight intensities and directions through real-time or pre-baked ray-casting. This allows for the rendering of horizon gradients, as well as the reflection, refraction, diffusion, scattering, and dispersion of light. The weather phenomena are procedurally generated using volumetric effects. These include static/dynamic clouds, fog, mist, precipitation (rain and snow), and stochastic wind gusts. The time and weather conditions can be continuously updated or set to one of the high-level presets (e.g., sunny, cloudy, foggy, rainy, snowy, etc.).

\subsubsection{Application Programming Interface}
\label{Sub-Sub-Section: Application Programming Interface}

In order to facilitate bi-directional data flow between the digital twins and the system under test (SUT), an application programming interface (API) is necessary. Consequently, a significant effort went into developing an object-oriented WebSocket protocol to establish a direct, real-time communication bridge between AutoDRIVE and MathWorks\textsuperscript{\textregistered} tools. The base API is directly compatible with MATLAB\textsuperscript{\textregistered} scripts, and can be wrapped as Level-2 MATLAB\textsuperscript{\textregistered} S-Functions to create custom blocks that seamlessly integrate the API with Simulink\textsuperscript{\texttrademark} models. The said API can be abstracted and exposed to enable granular control of the vehicles (sensor data streaming, actuator/light control, etc.) and their operating environments (time of day, weather, etc.).

\subsection{Integrated Model-Based Engineering}
\label{Sub-Section: Integrated Model-Based Engineering}

Integrated model-based engineering (IMBE) is a digital engineering paradigm that combines the power of MBSE and MBD into a cohesive framework. Here, MBSE focuses on defining, analyzing, and managing the system architecture, requirements, and behaviors, while MBD focuses on developing, simulating, and testing the system. This enables a more integrated, collaborative, and efficient approach to SUT development and validation\footnote{\url{http://github.com/AutoDRIVE-Ecosystem/AutoDRIVE-MathWorks}}.

\subsubsection{System Architecture}
\label{Sub-Sub-Section: System Architecture}

The high-level conceptual designs of a system can be captured within an architecture model, which includes the specifications, compositions, and interfaces between the different components and sub-systems. The proposed framework leverages System Composer\textsuperscript{\texttrademark} to define a hierarchical system architecture of the off-road autonomy algorithm. System Composer\textsuperscript{\texttrademark} offers integrated support for a variety of modeling, importing, and exporting formats, facilitating a seamless and collaborative architecture authoring experience.

\begin{figure}[t]
\centering
\includegraphics[width=\linewidth]{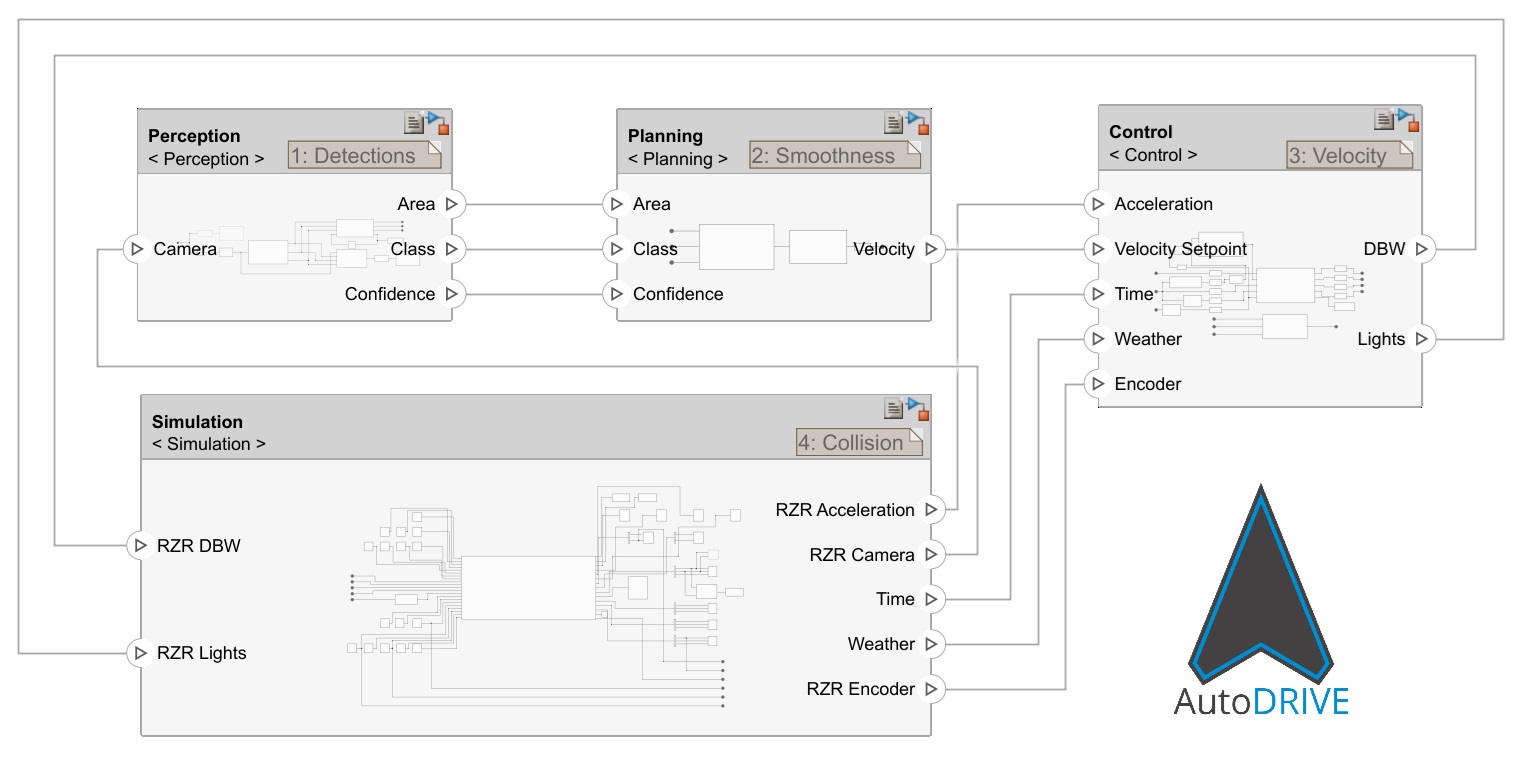}
\caption{High-level system architecture depicting the perception, planning, and control sub-systems of the autonomy algorithm along with the simulation component, all linked with requirements for complete traceability.}
\label{fig5}
\end{figure}

\subsubsection{Requirements Engineering}
\label{Sub-Sub-Section: Requirements Engineering}

\begin{table*}[t]
\caption{Defining and linking hierarchical requirements of the SUT.}
\begin{center}
\resizebox{\textwidth}{!}{%
\begin{tabular}{l|l|l|l}
\hline
\multicolumn{1}{c|}{\textbf{\begin{tabular}[c]{@{}c@{}}ID\end{tabular}}} &
    \multicolumn{1}{c|}{\textbf{\begin{tabular}[c]{@{}c@{}}SUMMARY\end{tabular}}} &
    \multicolumn{1}{c|}{\textbf{\begin{tabular}[c]{@{}c@{}}DESCRIPTION\end{tabular}}} &
    \multicolumn{1}{c}{\textbf{\begin{tabular}[c]{@{}c@{}}LINKS\end{tabular}}} \\ \hline
R1   &  Detection   &  \begin{tabular}[c]{@{}l@{}}The perception sub-system shall be able to detect various objects on the road ahead, \\classify them, and provide the confidence scores of the detections. That is to say that \\the number of detections throughout the test shall be greater than 1.\end{tabular}         &  \begin{tabular}[c]{@{}l@{}}Implemented by C1\\Verified by V1\end{tabular} \\ \hline
R2   &  Comfort   &   \begin{tabular}[c]{@{}l@{}}The planning sub-system shall ensure passenger comfort while ensuring effectiveness \\in stopping the vehicle. That is to say that the peak jerk experienced by the RZR \\shall be less than 6 m/s$^3$.\end{tabular}         & \begin{tabular}[c]{@{}l@{}}Implemented by C2\\Verified by V2\end{tabular} \\ \hline
R3   &  Tracking     &   \begin{tabular}[c]{@{}l@{}}The control sub-system shall accurately track the prescribed velocity setpoint, with \\minimal deviation. That is to say that the average estimated velocity error shall be \\within acceptable bounds of $\pm$ 1 m/s.\end{tabular}         & \begin{tabular}[c]{@{}l@{}}Implemented by C3\\Verified by V3\end{tabular} \\ \hline
R4   &  Safety    &   \begin{tabular}[c]{@{}l@{}}The system shall prevent any collision under all valid scenarios within its operational \\limits. That is to say that the total number of collisions of the RZR shall be 0.\end{tabular}       & \begin{tabular}[c]{@{}l@{}}Implemented by C4\\Verified by V4\end{tabular} \\ \hline
\end{tabular}
}
\end{center}
\label{tab1}
\end{table*}

The system architecture can be enriched by specifying the requirement(s) at the component, sub-system, or system levels. System Composer\textsuperscript{\texttrademark} allows authoring, managing, annotating, and linking requirements within the system architecture.

\subsubsection{Model-Based Design}
\label{Sub-Sub-Section: Model-Based Design}

MBD allows the development of the SUT as a digital model. Particularly, the components and sub-systems of the SUT can be implemented as Simulink\textsuperscript{\texttrademark} models within the system architecture abstraction layer of System Composer\textsuperscript{\texttrademark} (refer the inset in Fig. \ref{fig1}). This powerful tool also allows for efficient variant management.

\subsubsection{Test-Case Definition}
\label{Sub-Sub-Section: Test-Case Definition}

The proposed framework lies can facilitate efficient variant management, and granular parameter sweeps, allowing for the generation of a comprehensive bank of unique test cases. The framework supports both manual and scripted iterations, to define the number and sequence of test cases to be executed.

\subsubsection{Automated Test Execution}
\label{Sub-Sub-Section: Automated Test Execution}

Once the test cases are defined and the termination criteria are set, an automated test manager schedules the test execution. Each test is automatically executed by compiling the SUT variant and spinning up a digital twin simulation instance, followed by running the test case until the specified termination criteria are met, and finally gracefully terminating the test case by logging the data, clearing the cache, and killing the simulation instance. The test manager performs these operations recursively until the test bank is exhausted. Throughout the execution, the test manager keeps track of the test cases that are running, pending, and completed.

\subsubsection{Data Logging and Storage}
\label{Sub-Sub-Section: Data Logging and Storage}

The rich data generated from each test execution is continuously logged for post-processing and analysis. This data can include key performance indicators (KPIs), test results as well as other metadata related to the individual test cases. Apart from the raw data, System Composer\textsuperscript{\texttrademark} also supports automated report generation for user convenience.


\section{Case Study}
\label{Section: Case Study}

This section elucidates the off-road navigation case study of an autonomous RZR performing visual servoing to drive along a dirt road and react to any obstacles or environmental changes. The main objective of this case study was to create a simple and intuitive example to demonstrate the efficacy of the proposed framework, although it has the potential to handle much more complex system-of-systems level designs. Here, we first describe the high-level use case before delving into the details with a correlation to the proposed digital engineering framework presented earlier.

\subsubsection{System Under Test}
\label{Sub-Sub-Section: System Under Test}

A jump-scare scenario was devised to analyze the ego vehicle's reactive response to a \textit{``panic''} event. Particularly, the ego vehicle was supposed to perform visual servoing to drive along a dirt road, unless it was unsafe/unethical to do so (e.g., dead-end, imminent collision, etc.). Down the road, the ego vehicle would encounter a herd of animals (e.g., deer, moose, caribou, etc.) blocking the drivable area, requiring the ego vehicle to execute an emergency braking maneuver (refer Fig. \ref{fig1}).

\subsubsection{System Architecture}
\label{Sub-Sub-Section: CS: System Architecture}

The concept-level system architecture comprises perception (C1), planning (C2), and control (C3) sub-systems of the SUT, along with the simulation API module (C4) (refer Fig. \ref{fig5}).

\subsubsection{Requirements Engineering}
\label{Sub-Sub-Section: CS: Requirements Engineering}

Table \ref{tab1} hosts the requirements imposed on the perception (R1), planning (R2), and control (R3) sub-systems, as well as those imposed on the system (R4) as a whole.

\subsubsection{Model-Based Design}
\label{Sub-Sub-Section: CS: Model-Based Design}

We first implemented C4, which provides sensor streams including drive-by-wire (DBW) feedback \{$\tau$, $\delta$, $\chi$, $\xi$\}, encoder data \{$N_{\text{ticks}}$, $N_{\text{rev}}$\}, INS data \{$x$, $y$, $z$, $\phi$, $\theta$, $\psi$, $\ddot{x}$, $\ddot{y}$, $\ddot{z}$, $\dot{\phi}$, $\dot{\theta}$, $\dot{\psi}$\}, camera frames \{\texttt{cam\_fl}, \texttt{cam\_fr} $\in \mathbb{R}^{1280 \times 720 \times 3}$\}, and LIDAR PCD \{$\mathbf{x_{\text{hit}}, y_{\text{hit}}, z_{\text{hit}}}$\}, as well as ground-truth data such as the distance to collision \{\texttt{dtc}\} and collision count \{\texttt{n\_col}\}. This sub-system also accepts environment controls including time of day \{\texttt{tod}\} and weather conditions \{\texttt{weather}\}, as well as vehicle controls including DBW commands \{$\tau_\text{cmd}$, $\delta_\text{cmd}$, $\chi_\text{cmd}$, $\xi_\text{cmd}$\}, light commands \{headlights, DRL\}, and co-simulation updates (if applicable).

C1 is implemented to accept \texttt{cam\_fr} frame, down-sample it to $\mathbb{R}^{640 \times 360 \times 3}$, and feed it to a deep-learning object detector model, which returns detections in the form of bounding boxes, classification labels, and confidence scores. The detection count \{\texttt{n\_det}\} is updated based on the number of detections, and the detections are post-processed to compute the \texttt{class}, \texttt{size}, and \texttt{confidence} of the detected objects. Here, we define 2 variants of C1, viz. C1.1 and C1.2, which use distinct deep learning models. Particularly, C1.1 implements \texttt{tiny-yolo-v2} detector, and C1.2 implements \texttt{tiny-yolo-v3} detector.

C2 is implemented to filter the detections based on their \texttt{class} (we pool together 10 labels from the COCO dataset \cite{coco-dataset-2014}, which are representative of the \textit{``animal''} class), \texttt{size} ($\geq$2500 px$^2$), and \texttt{confidence} ($\geq$50\%). This information is used to generate an autonomous emergency braking (AEB) trigger \{\texttt{aeb} $\in [0, 1]$\}, which is then passed on to a velocity profiler to determine the vehicle's reference velocity:
$
v_{ref} = \begin{cases}
0 & \text{if } \texttt{aeb} \geq 0.9 \\
\frac{0.3}{\texttt{aeb}+0.1} & \text{otherwise}
\end{cases}
$.
Here, we define 2 variants of C2, viz. C2.1 and C2.2, which use distinct AEB triggers. Particularly, C2.1 implements an abrupt AEB trigger, i.e., \texttt{aeb} $\in \{0, 1\}$, while C2.2 implements a smooth AEB trigger, i.e., $\texttt{aeb} = \texttt{max}(0, \texttt{min}(1, 1e-4*\texttt{size}))$.

C3 is implemented to track the velocity setpoint using a feedback control law. The velocity feedback is estimated by fusing IMU and encoder data. Particularly, the longitudinal accelerations are numerically integrated to compute one of the velocity estimates $^av_t = \:^av_{t-1} + \frac{a_t + a_{t-1}}{2} * \Delta t$, while the encoder ticks are numerically differentiated to compute the other: $^ev_t=\text{max}\left(-30, \text{min}\left(30, \frac{N_{\text{ticks}_t}-N_{\text{ticks}_{t-1}}}{\Delta t}\right)\right) * \frac{2*\pi*r_{\text{wheel}}}{PPR}$. Here, we define 2 variants of C3, viz. C3.1 and C3.2, which use distinct feedback controllers. Particularly, C3.1 implements a bang-bang controller, i.e.,
$
\begin{cases}
\tau = 0.4 & \text{if } \epsilon_v > 0 \\
\chi = 0.4 & \text{otherwise}
\end{cases}
$,
while C3.2 implements a bounded PID controller, i.e., $\tau = \texttt{max}(0, \texttt{min}(0.5, \texttt{PID($\epsilon_v$)}))$. C3 also incorporates hill-hold assist control (HAC), which uses the encoder rates $\left(\frac{N_{\text{ticks}_t}-N_{\text{ticks}_{t-1}}}{\Delta t}\right)$ to estimate back-rolling due to the inclined terrain and engages brakes to avoid it. Additionally, C3 implements adaptive light control (ALC) to switch vehicle lights (headlights and DRL) based on the time of the day and weather conditions.

\subsubsection{Test Matrix}
\label{Sub-Sub-Section: CS: Test Matrix}

This work defines 8 variant configurations comprising 2 variants each for perception \{C1.1, C1.2\}, planning \{C2.1, C2.2\}, and control \{C3.1, C3.2\}) sub-systems, along with 16 parameter sets comprising 4 times of the day (\{P1.1 = 10:00 am, P1.2 = 01:00 pm, P1.3 = 04:00 pm, P1.4 = 12:00 am\} and 4 weather presets \{P2.1 = clear, P2.2 = fog, P2.3 = rain, P2.4 = snow\}). The 8 variant configurations are crossed with the 16 parameter sets (\{C1.1, C1.2\} $\times$ \{C2.1, C2.2\} $\times$ \{C3.1, C3.2\}) $\times$ (\{P1.1, P1.2, P1.3, P1.4\} $\times$ \{P2.1, P2.2, P2.3, P2.4\}) to generate a battery of 128 test cases.

\subsubsection{Test Results}
\label{Sub-Sub-Section: CS: Test Results}

\begin{table*}[t]
\caption{Normalized verification scores of the SUT validated across the 128-test matrix.}
\begin{center}
\resizebox{\textwidth}{!}{%
\begin{tabular}{l|llllll|llllllll|l}
\hline
\multicolumn{1}{c|}{\multirow{2}{*}{\textbf{TEST}}} & \multicolumn{6}{c|}{\textbf{VARIANTS}}                                                                                                                                                                                   & \multicolumn{8}{c|}{\textbf{PARAMETERS}}                                                                                                                                                                                                                                                        & \multicolumn{1}{c}{\multirow{2}{*}{\textbf{TOTAL}}} \\ \cline{2-15}
\multicolumn{1}{c|}{}                               & \multicolumn{1}{c}{\textbf{C1.1}} & \multicolumn{1}{c|}{\textbf{C1.2}} & \multicolumn{1}{c}{\textbf{C2.1}} & \multicolumn{1}{c|}{\textbf{C2.2}} & \multicolumn{1}{c}{\textbf{C3.1}} & \multicolumn{1}{c|}{\textbf{C3.2}} & \multicolumn{1}{c}{\textbf{P1.1}} & \multicolumn{1}{c}{\textbf{P1.2}} & \multicolumn{1}{c}{\textbf{P1.3}} & \multicolumn{1}{c|}{\textbf{P1.4}} & \multicolumn{1}{c}{\textbf{P2.1}} & \multicolumn{1}{c}{\textbf{P2.2}} & \multicolumn{1}{c}{\textbf{P2.3}} & \multicolumn{1}{c|}{\textbf{P2.4}} & \multicolumn{1}{c}{}                                  \\ \hline
\textbf{V1}                                         & 0.6563                            & \multicolumn{1}{l|}{\textbf{0.9844}}        & 0.8125                            & \multicolumn{1}{l|}{\textbf{0.8281}}        & 0.7969                            & \textbf{0.8438}                             & \textbf{1.0000}                            & \textbf{1.0000}                            & 0.5000                            & \multicolumn{1}{l|}{0.7813}        & 0.7813                            & \textbf{0.8438}                            & 0.8125                            & \textbf{0.8438}                             & \textbf{0.8203}                                                \\
\textbf{V2}                                         & 0.3906                            & \multicolumn{1}{l|}{\textbf{0.4219}}        & \textbf{0.4219}                            & \multicolumn{1}{l|}{0.3906}        & 0.0156                            & \textbf{0.7969}                             & 0.3438                            & 0.3125                            & \textbf{0.5000}                            & \multicolumn{1}{l|}{0.4688}        & \textbf{0.5000}                            & 0.3750                            & 0.3438                            & 0.4063                             & 0.4063                                                \\
\textbf{V3}                                         & \textbf{0.2188}                            & \multicolumn{1}{l|}{0.1875}        & \textbf{0.2031}                            & \multicolumn{1}{l|}{\textbf{0.2031}}        & 0.0625                            & \textbf{0.3438}                             & \textbf{0.4375}                            & 0.1563                            & 0.0938                            & \multicolumn{1}{l|}{0.1250}        & 0.1875                            & 0.1875                            & 0.1875                            & \textbf{0.2500}                             & 0.2031                                                \\
\textbf{V4}                                         & \textbf{0.4844}                            & \multicolumn{1}{l|}{0.4688}        & \textbf{0.5625}                            & \multicolumn{1}{l|}{0.4063}        & \textbf{0.6094}                            & 0.3438                             & \textbf{0.8125}                            & 0.6875                            & 0.1875                            & \multicolumn{1}{l|}{0.2188}        & 0.4688                            & 0.4375                            & \textbf{0.5000}                            & \textbf{0.5000}                             & 0.4766                                                \\ \hline
\textbf{All}                                        & 0.0781                            & \multicolumn{1}{l|}{\textbf{0.0938}}        & \textbf{0.1250}                            & \multicolumn{1}{l|}{0.0469}        & 0.0156                            & \textbf{0.1563}                             & \textbf{0.2813}                            & 0.0625                            & 0.0000                            & \multicolumn{1}{l|}{0.0000}        & \textbf{0.0938}                            & \textbf{0.0938}                            & \textbf{0.0938}                            & 0.0625                             & \textbf{0.0859}                                                \\ \hline
\end{tabular}
}
\end{center}
\label{tab2}
\end{table*}

\begin{figure}[t]
\centering
\includegraphics[width=\linewidth]{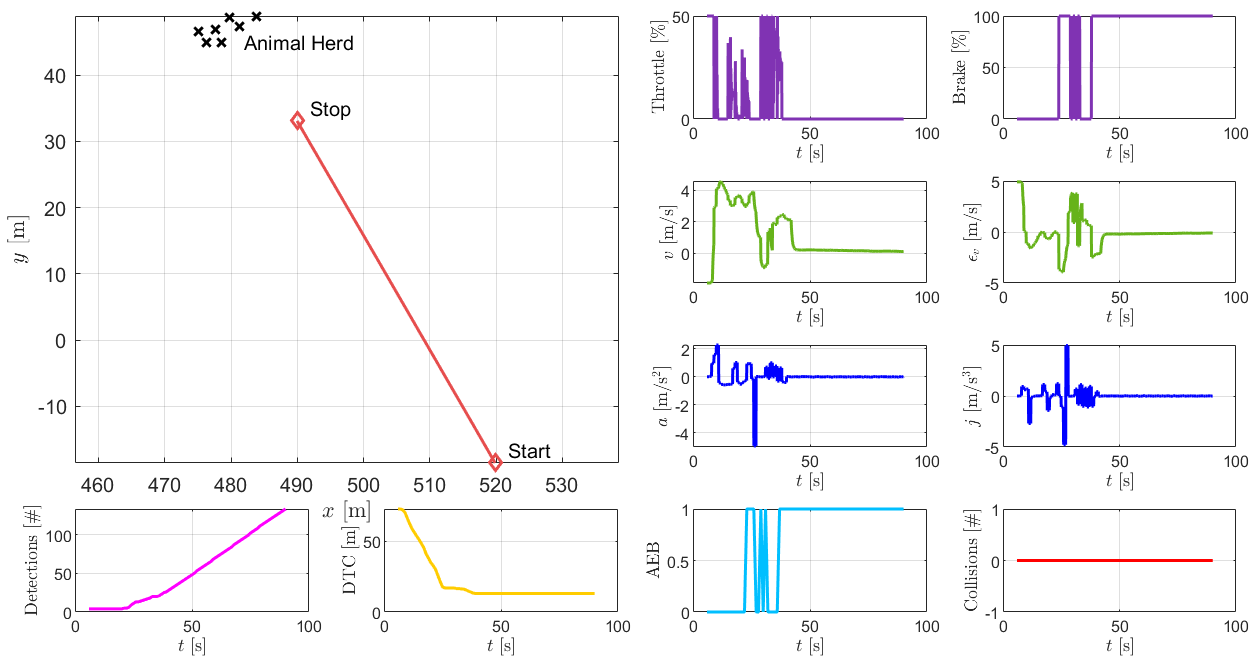}
\caption{Automated data analysis performed over the top 12 KPIs recorded from the 15$^\text{th}$ test iteration \{C1.2, C2.1, C3.2, P1.1, P2.2\}.}
\label{fig6}
\end{figure}

After executing all the tests, the data logged from each test was automatically post-processed and analyzed. The said KPIs and test results were also automatically exported as a PDF report with hyperlinks for traceability.

Fig \ref{fig6} depicts data visualization for one of the test cases, highlighting select KPIs including those used to verify the requirement satisfaction. The positional plot highlights the odometry of the vehicle along the dirt road and assesses its distance from the animal herd. This is also reflected in the DTC plot, which shows that the vehicle came to a safe stop $\sim$13 m from the animal herd. The zero collision count denotes that the SUT has passed its system-level requirement (R4), and we can similarly analyze the sub-system-level requirement satisfaction. The number of object detections has accumulated exponentially to $\sim$130 during the 90-second test, which highlights that R1 has been satisfied. Additionally, the zero detection count until the 20-second mark is useful in establishing the reliable range of the object detector. The estimated velocity, acceleration, and jerk values capture the motion profile of the vehicle. It can be observed that the vehicle reached a peak velocity of 4.6 m/s, with a peak acceleration of 2.26 m/s$^2$. It experienced a maximum deceleration of -4.9 m/s$^2$ around the 26-second mark when the AEB was first triggered. This resulted in a peak jerk of 4.94 m/s$^3$, which was within the prescribed limits ($\pm$6 m/s$^3$) to satisfy R2. Finally, the performance of the control sub-system can be assessed by analyzing the control actions relayed to the vehicle, which do not show any conflicts. Additionally, the tracking error profile highlights the control performance, and its mean value (-0.11 m/s) lies within the prescribed limits ($\pm$1 m/s) to satisfy R3.

Table \ref{tab2} hosts the normalized verification scores of the SUT, for the entire 128 test-matrix. These scores represent the fraction of test cases passed out of the total test cases for each variant configuration and parameter set. The rows depict requirement verification at the individual level (V1 to V4) and the collective level (All = V1 $\wedge$ V2 $\wedge$ v3 $\wedge$ V4). The last column denotes requirement satisfaction across all the tests -- we can observe that R1 has been satisfied the most, followed by R4 and R2, while R3 has been satisfied the least. This highlights the fact that while primary objectives (detection, safety) were met in most cases, secondary objectives (comfort, tracking) were sometimes compromised.

We further demonstrate how verification scores across different variants can quickly inform design choices with credibility. For the perception sub-system, we can observe that the R1 satisfaction rate of C1.2 is significantly higher than that of C1.1. Furthermore, its overall satisfaction rate is also higher, making it a better design choice. Similarly, for the planning sub-system, although the R2 satisfaction rate of C2.1 and C2.2 is almost the same, the overall satisfaction with C2.1 is over two times higher, giving it a competitive edge in terms of reliability. Finally, for the control sub-system, we can see that C3.2 dominates across all requirements, except for R4. Consequently, it can be established that while C3.2 is better in terms of tracking and comfort, C3.1 is much better in terms of safety.

Finally, the verification scores across different parameters can help reliably define the operational design domain of the SUT. We can observe that the SUT did not perform well in the evening/night, which can be primarily attributed to the perception sub-system. None of the test cases during these times passed all the requirements. Consequently, the SUT may target only morning and afternoon times of the day, which highlight perfect scores for R1. A similar comparison among various weather conditions reveals that the SUT performs equally across all of them, with slightly less reliability during snowfall. Consequently, the SUT may target all weather conditions, with extra caution when it is snowing.


\section{Conclusion}
\label{Section: Conclusion}

In this paper, we presented a comprehensive digital engineering framework to enable the systematic development and validation of autonomy algorithms for off-road ground vehicles. Particularly, this work focused on integrating digital twin simulations with MBSE and MBD workflows to formulate a cohesive framework that not only improves the efficiency of the development process but also ensures rigorous verification and validation. We substantiated the framework's efficacy by demonstrating the successful development and validation of a candidate SUT involving an autonomous LTV performing visual servoing to navigate a dirt road to help support our claims. By considering various test parameters such as times of the day and weather conditions, along with different algorithmic variants across the perception, planning, and control sub-systems, we elucidated how the proposed framework ensures a thorough evaluation of the autonomous vehicle's performance in a wide range of scenarios automatically.

Future research could delve into the investigation of other and potentially more complex SUTs, as well as employing this framework to autonomous vehicles across different scales and operational design domains. Additionally, the use of high-performance computing (HPC) resources and hardware-in-the-loop (HiL) testbeds could enhance this framework.


\begin{ack}
This work was supported by the Virtual Prototyping of Autonomy Enabled Ground Systems (VIPR-GS), a U.S. Army Center of Excellence for modeling and simulation of ground vehicles, under Cooperative Agreement W56HZV-21-2-0001 with the U.S. Army DEVCOM Ground Vehicle Systems Center (GVSC).
\end{ack}


\balance
\bibliography{ifacconf}


\end{document}